\pdfoutput=1

\documentclass[11pt]{article}

\usepackage[preprint]{acl}

\usepackage{times}
\usepackage{latexsym}

\usepackage[T1]{fontenc}

\usepackage[utf8]{inputenc}

\usepackage{microtype}

\usepackage{inconsolata}

\usepackage{graphicx}
\usepackage[nameinlink]{cleveref}
\usepackage{color,soul}
\usepackage{multirow}
\usepackage{booktabs}
\usepackage{multirow}
\usepackage{comment}
\title{CULL-MT: Compression Using Language and Layer pruning\\ for Machine Translation}

\author{Pedram Rostami \\
  University of Tehran \\
\texttt{pedram.rostami@ut.ac.ir} \\\And
  Mohammad Javad Dousti \\
  University of Tehran \\
\texttt{mjdousti@ut.ac.ir} \\}

\begin{document}
\maketitle
\begin{abstract}
Multilingual machine translation models often outperform traditional bilingual models by leveraging translation knowledge transfer. 
Recent advancements have led to these models supporting hundreds of languages and achieving state-of-the-art results across various translation directions. 
However, as these models grow larger, their inference operations become increasingly costly.
In many use cases, there is no need to support such a wide range of language pairs, as translation is typically needed in only a few selected directions. 
In this paper, we present CULL-MT, a compression method for machine translation models based on structural layer pruning and selected language directions.
Our approach identifies and prunes unimportant layers using a greedy strategy, then mitigates the impact by applying knowledge distillation from the original model along with parameter-efficient fine-tuning.
We apply CULL-MT to the NLLB-3.3B and LLaMA3.1-8B-Instruct models. 
In a multi-way translation scenario (Persian, French, and German to English), we find the NLLB-3.3B model to be robust, allowing 25\% of layers to be pruned with only a 0.9 spBLEU drop. 
However, LLaMA3.1-8B-Instruct is more sensitive, with a 2.0 spBLEU drop after pruning 5 layers.


  
  

\end{abstract}

\section{Introduction}

In recent years, multilingual neural machine translation (NMT) models have become an efficient solution for translating multiple language pairs, compared to using separate models for each translation direction
\cite{aharoni-etal-2019-massively}.
Additionally, multilingual NMT models \cite{firat-etal-2016-multi} outperform bilingual NMT models \cite{bahdanau-2015-nmt}, particularly in low-resource language pairs, due to the transfer of knowledge from high-resource language pairs
\cite{dabre-etal-2020-survey}
.
Therefore, recent multilingual NMT models have increasingly focused on expanding the number of supported languages. 
Notable models like NLLB \cite{costa2022no} and MADLAD \cite{madlad400} now support 200 and 400 languages, respectively, with many of these being low-resource languages.


Despite their impressive performance, multilingual NMT models have grown significantly in size, often containing billions of parameters.
This increase has made their deployment and inference computationally expensive, requiring substantial hardware resources. 
%
The parameter count of multilingual NMT models, such as NLLB \cite{costa2022no} and MADLAD \cite{madlad400}, has surged in recent years, mirroring the trend in language models, which gave rise to the term \textit{Large Language Models} (LLMs). 
This rapid expansion raises concerns about the cost and practicality of using these models in real-world applications, particularly in resource-constrained environments. 
As a result, compressing large models without significantly impacting performance has become a critical research challenge.

While recent multilingual NMT models support hundreds of languages, many medium- and small-sized translation systems require only a few translation directions. 
Compressing massive translation models to focus on the relevant language pairs can result in significant computational savings. 
Our approach addresses this need by compressing the model with minimal performance sacrifice in the selected translation tasks.

Both multilingual NMT models and large language models are designed to handle a wide range of tasks. 
While multilingual NMT models can translate hundreds of language directions, LLMs are versatile enough for various natural language processing (NLP) tasks, including multilingual translation \cite{zhao2023survey}. 
Our approach focuses on compressing these models for specific tasks. 
For multilingual NMT, we reduce model size while maintaining performance on key translation directions. 
Similarly, for LLMs, we compress the model specifically for multilingual translation of selected language pairs, showcasing the versatility and architecture-agnostic nature of our method.

For addressing this issue, we introduce CULL-MT, a method that compresses models by focusing on the key translation directions relevant to specific use cases, achieving notable computational savings with minimal performance sacrifice.
Our method employs a greedy structural pruning technique, wherein we iteratively assess the importance of each layer by comparing the model's performance with and without that layer. 
If removing a layer results in only a minimal performance drop, it is considered less critical and is subsequently pruned. 
This iterative process continues until the performance falls below a predefined threshold. 
After the pruning phase, the model is fine-tuned to recover any performance loss, ensuring effectiveness for the targeted translations.

The following key contributions highlight the
effectiveness of our CULL-MT approach:

We applied our approach to the NLLB-3.3B model, demonstrating that it is considerably robust against layer pruning. 
In a multi-way scenario (translating from Persian, French, and German to English), we were able to prune 12 out of 48 layers, with the pruned model experiencing only a 0.9 drop in spBLEU score. 
In a single-way scenario (translating from English to Persian), CULL-MT allowed us to prune 15 layers while resulting in a 1.2 drop in spBLEU score.

Our approach was also tested on the LLaMA3.1-8B-Instruct model, revealing its sensitivity to layer pruning in translation tasks. 
In the multi-way scenario, CULL-MT pruned 5 layers, resulting in a 2.0 drop in spBLEU score. 

Analyzing the importance of layers in the multi-way translation task for both NLLB-3.3B and LLaMA3.1-8B-Instruct revealed critical insights. 
Our findings indicate that the first layer in the encoder and the decoder of the NLLB-3.3B model are essential, while other layers have a considerably lesser impact on performance. 
In contrast, the LLaMA3.1-8B-Instruct model exhibits more significant important regions, with the first two and last five layers being crucial, alongside several important layers in the middle of the model.

The remainder of this paper is organized as follows: 
\Cref{sec:related} reviews common compression methods and layer-wise pruning techniques. 
\Cref{sec:cull_mt} presents the CULL-MT methodology, while \Cref{sec:experiments} outlines the experimental setup. 
\Cref{sec:results} reports the results of CULL-MT and compares them with other approaches. 
\Cref{sec:layer_importance} provides an analysis of layer importance, and \Cref{sec:conclusion} concludes the paper. 
Finally, \Cref{sec:limitations} discusses the limitations.



    
    

\begin{figure*}[h]
  \centering
  \includegraphics[width=0.9\textwidth]{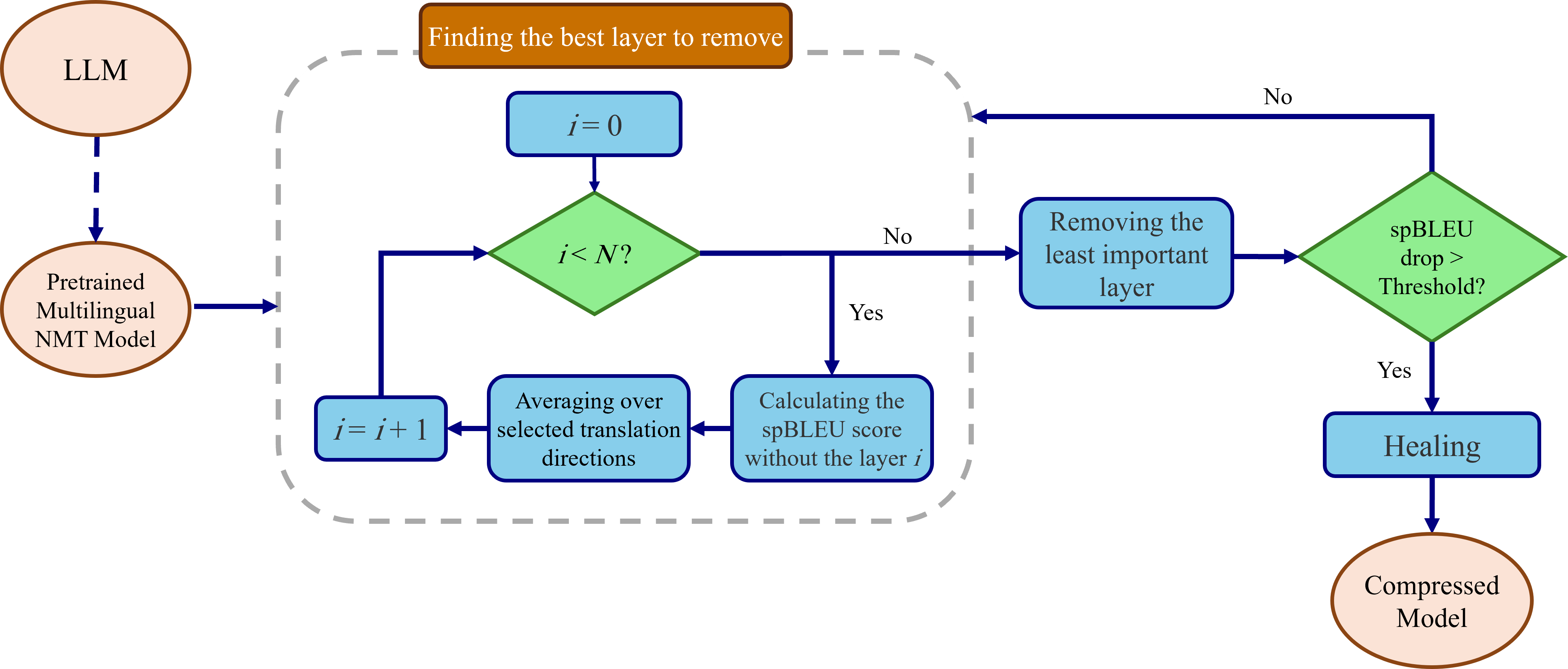}
  \caption {Overview of the CULL-MT approach. It iteratively prunes a multilingual translation model by identifying and removing the least important layers while preserving performance in key translation directions.}
  \label{fig:approach}
\end{figure*}
  %

\section{Related Works}
\label{sec:related}
\subsection{Compressing Massive Models}

The rapid increase in the size of pre-trained language models \cite{devlin-etal-2019-bert}, such as GPT-3 \cite{Brown-2020-gpt3} and LLaMA \cite{touvron2023llama}, has led to substantial interest in model compression techniques to reduce inference costs and memory footprints. 
Common methods include quantization \cite{jacob-2018-quantization}, which reduces precision in model weights, and knowledge distillation \cite{hinton-2015-distilling}, which transfers knowledge from a larger model to a smaller one while retaining performance on downstream tasks. 
Pruning, another effective method, involves removing unimportant weights or neurons from a model \cite{han-2015-pruning} to achieve a more compact version without significant performance degradation.

In addition to advancements in model compression, significant progress has been made in parameter-efficient fine-tuning (PEFT) methods, which facilitate the adaptation of LLMs for consumer-level devices.
These techniques allow for effective fine-tuning while minimizing the computational burden \cite{lialin2023scaling}.
One prominent example is Low-Rank Adaptation (LoRA), which utilizes low-rank matrices to enhance fine-tuning efficiency \cite{hu2022lora}. 
Furthermore, some approaches have combined various compression and PEFT methods to maximize efficiency.
For example, QLoRA employs 4-bit double quantization of the original model alongside LoRA weights, facilitating low-resource fine-tuning \cite{dettmers2023qlora}.
Moreover, Minitron utilizes a combination of pruning and knowledge distillation to create smaller, efficient language models from larger pretrained architectures \cite{muralidharan2024compact}.

\subsection{Structural Pruning}

Pruning \cite{lecunn-1989-optimal} is a key approach for compressing neural networks, aiming to identify and remove less important weights using various methods. 
%
While unstructured pruning allows for the sparsification of specific weights, it often necessitates specialized hardware for speed improvements \cite{cheng-2024-pruning-survey}.
In contrast, structural pruning \cite{molchanov2017pruning} focuses on removing entire filters, blocks, or layers, providing efficiency that is not dependent on specific devices \cite{cheng-2024-pruning-survey}.

\citet{voita-etal-2019-bottom} indicated that the top layers of pre-trained transformer \cite{vaswani-2017-transformer} models are often specialized for the original training objectives.
Therefore, \citet{sajjad-etal-2023-layers-effect} showed that the higher layers are not essential for fine-tuning on downstream tasks and they can be pruned.
\citet{peer-2022-glp} introduce a Greedy Layer Pruning (GLP) method that iteratively selects layers to remove, optimizing performance on GLUE tasks with encoder-only models. 
While their focus is on GLUE tasks, we concentrate on translation tasks involving encoder-decoder and decoder-only architectures.

Recent works concentrate on pruning LLMs.
\citet{ma2023llmpruner} developed LLM-Pruner, which uses structural pruning to remove non-critical coupled structures based on gradient information, effectively compressing large language models while maintaining their core functionality. 
This method allows for performance recovery with tuning techniques like LoRA and demonstrates satisfactory zero-shot task capabilities across various models.
Furthermore, \citet{gromov2024unreasonable} highlighted that deeper layers often exhibit highly similar representations. 
This similarity allows for significant model compression, particularly by pruning deeper layers starting from the penultimate layer, which can be achieved without major performance loss when combined with light fine-tuning.

    

\section{CULL-MT}
\label{sec:cull_mt}

Our approach focuses on structurally pruning a multilingual translation model by iteratively identifying and removing the least important layers, while preserving performance on key translation directions. 
The procedure involves two nested iterations: an inner loop for identifying the best layer to remove, and an outer loop that repeats this procedure until the model’s performance drop exceeds a predefined threshold. 
The process is visualized in \Cref{fig:approach}, which gives an overall view of our approach.

\textbf{Layer Importance Evaluation.}
The core of our methodology is evaluating the importance of each layer. 
For this, we measure the model’s performance with and without each layer using a development translation dataset. 
Specifically, we assess the spBLEU \cite{costa2022no, papineni-etal-2002-bleu} score for the selected translation directions. 
After testing each layer, we average the spBLEU scores across all directions of interest to determine a layer’s overall importance. 
The layer whose removal results in the smallest performance drop is considered the least important and is pruned.

\textbf{Iterative Layer Pruning.}
The inner loop of the algorithm --- referred to as "Finding the best layer to remove" --- iterates over the model’s layers and evaluates the importance of each layer as described above. 
Once all layers have been evaluated, we identify the best layer to prune, i.e., the one which minimally impacts the model’s performance.

After each pruning step, the outer loop repeats the inner procedure, re-evaluating the remaining layers. 
By doing so, our approach continually assesses how the removal of one layer affects the importance of other layers, ensuring that we prune in a way which maintains maximum performance for the selected directions.

By stopping the pruning process at a specific threshold and utilizing efficient fine-tuning techniques, running the CULL-MT approach on the NLLB-3.3B and LLaMA3.1-8B-Instruct models takes less than a day. 
This is not a significant concern, as the process is intended to run only once to create the compressed model.

Our greedy approach identifies local solutions for determining the optimal set of layers to prune. 
Although we explored other methods, such as dynamic programming, they did not yield better results. 
This is because the effect of removing a specific layer depends on the layers that have already been removed, making it difficult to generalize. 
Therefore, finding the truly optimal solution requires evaluating all possible combinations of layer removals, which is computationally infeasible due to the sheer number of configurations.

\textbf{Focus on Special Translation Directions.}
Our approach focuses on preserving performance for specific translation directions by evaluating layer importance based on the averaged performance across these directions. 
In real-world scenarios, where certain translation directions may hold greater importance, we can assign higher weights to those directions when calculating the spBLEU score drop during averaging. 
This allows the pruning process to prioritize key translation directions while still considering others. 
More generally, each direction can be assigned a custom weight based on its priority, providing flexibility to adapt the pruning process to real-world needs and focus on critical translation tasks.

\textbf{Healing.}
To restore any performance loss after pruning, we perform fine-tuning using sequence-level knowledge distillation \cite{kim-rush-2016-sequence} from the original model to the pruned model. 
Fine-tuning the pruned model with new datasets can result in catastrophic forgetting \cite{lange-2022-continual}. 
To address this, we utilize knowledge distillation from the original model, which helps maintain performance consistency and mitigate the risk of forgetting \cite{Li-2018-forgetting}.

To leverage sequence-level knowledge distillation, we use a generated dataset. 
The source sentences are drawn from the training datasets, while the target sentences are the translations produced by the original model. 
This generated dataset encompasses all the selected translation directions.

As recent multilingual NMT models and LLMs continue to grow in size, fine-tuning all their parameters has become increasingly challenging. 
Utilizing PEFT methods presents a viable solution \cite{lialin2023scaling}. 
Specifically, we employ the LoRA approach, which freezes the original model and injects trainable low-rank matrices into specific layers \cite{hu2022lora}. 
Additionally, LoRA is particularly effective with smaller training datasets, making it an ideal option for efficient fine-tuning in our context \cite{zhang2024when}.

    
    

\section{Experiments Setup}
\label{sec:experiments}

\subsection{Models}
In our experiments, we utilized two models. 
The first is the NLLB-3.3B model~\cite{costa2022no}, a multilingual NMT model supporting 200 languages, which follows an encoder-decoder architecture with 24 layers in both the encoder and decoder.
While NLLB is designed to handle a wide range of language directions, our compression approach focuses specifically on a subset of selected translation directions. 
We used the NLLB model as a fully pre-trained model without further fine-tuning.

The second model is the LLaMA3.1-8B model \cite{llama3.1, dubey2024llama}, an LLM known for its openness and strong performance across various NLP tasks.
This model follows a decoder-only architecture and contains 32 layers.
Although LLaMA3.1-8B-Instruct is versatile across various NLP tasks, we focused on compressing it specifically for the task of multilingual translation in certain language directions.
Before applying the compression, we performed instruction-tuning on it to enable zero-shot translation and improve the model's translation performance for this specific task across translation directions of interest.

\subsection{Datasets}

The training dataset is essential for enhancing the performance of both models in specific translation tasks, and we use the NLLB dataset \cite{costa2022no} as the source for our training data. 
For the NLLB-3.3B model, the healing procedure utilizes a training dataset consisting of 160,000 parallel sentences for each selected direction, which is critical for effectively refining the model. 
We conducted experiments on various training data sizes to determine that 160,000 sentences is optimal for our needs, as detailed in \Cref{sec:appendix1}.
In contrast, recent studies have shown that large language models do not require extensive datasets for fine-tuning in machine translation tasks \cite{zhang-etal-2023-machine}.
Consequently, we utilize the same 15,000 parallel sentences for each selected direction in the LLaMA3.1-8B-Instruct model for both the instruction-tuning and healing processes.

The \texttt{dev} dataset plays a critical role in the evaluation process, where it is employed to calculate the importance of each layer during the pruning procedure. 
For this purpose, we used the development subset of \textsc{Flores-200} dataset~\cite{costa2022no}, which provides high-quality and reliable multilingual translation data.

To evaluate the final pruned model and compare it against the original, we utilized the \texttt{test} subsets from Flores-200.
Additionally, for some developmental experiments, we used the \texttt{test} subset of the NTREX dataset \cite{federmann-etal-2022-ntrex}.

\subsection{Data Preprocessing and Prompt Format}
We did not perform any special data preprocessing for the development and test datasets, because they are already have high quality. 
The training data was sourced from the NLLB dataset, which has been preprocessed by the NLLB team and sorted based on cosine similarity scores between source and target sentences using the LASER3 model \cite{heffernan-etal-2022-bitext}. 
However, we observed that very short phrases or sentences tended to be located at the beginning of the dataset. 
To address this, we selected sentences which contained more words than the average for each specific language within the dataset. 
This approach allowed us to maintain a focus on more meaningful and informative translations during the training process.

For instruction tuning and healing of the LLaMA3.1-8B-Instruct model, we used the default template of the model to ensure compatibility with the original model. 
Additionally, we adopted the prompt format described by \citet{zhang-etal-2023-machine} to establish a clear boundary around translated sentences. 
For the system prompt, we utilized the following message: \texttt{You are a translator who translates messages from \{source language\} to \{target language\}} \cite{giray2023prompt}.

\subsection{Efficient Fine-tuning}
We fine-tune the models at different stages to optimize their performance. 
For LLMs, we initially fine-tune them to improve their zero-shot performance on the translation task.
After pruning layers from the models, we fine-tune them again to \textit{heal} the damage caused by removing layers. 
However, if the model already exhibits excellent performance in translation, there is no need for initial fine-tuning. 
Therefore, for multilingual NMT models such as NLLB-3.3B, we only perform fine-tuning in the healing phase.

For our experiments, we used an NVIDIA RTX 3090 with 24~GB of memory.
To fit a relatively model such as LLaMA3.1 (with 8 billion parameters) inside the GPU's memory, we utilized the \texttt{bfloat16} data type, which uses two bytes per parameter instead of four, thereby halving the memory requirement \cite{wang2019bfloat16}.
Additionally, \textit{gradient checkpointing} was applied to save memory by recomputing intermediate activations during the backward pass instead of storing them in memory \cite{chen2016training}.

We also employed ZeRO-Offload, which offloads optimizer states from the GPU to the CPU, allowing us to free up valuable GPU memory for other tasks \cite{ren2021zero}. 
While offloading increased training time, it enabled us to increase the batch size, which positively impacted the overall fine-tuning process. 
To further handle larger batch sizes, we employed \textit{gradient accumulation}, which splits the gradient calculation over multiple forward and backward passes, reducing the memory requirements for each pass.

Using these techniques, the memory consumption during fine-tuning is nearly equivalent to that of inference, ensuring efficient use of GPU resources.
While these methods allowed us to fine-tune all parameters of both models, we opted to use the LoRA technique. 
As will be shown in \Cref{sec:appendix2}, when the training data is limited, LoRA can outperform full parameter fine-tuning in terms of both efficiency and translation performance.

For the initial fine-tuning of the LLaMA3.1-8B-Instruct model, we utilized the optimal LoRA hyperparameters identified by \citet{zhang-etal-2023-machine}. 
However, during the healing process, we adjusted the LoRA rank to 128 and the LoRA $\alpha$ to 256 to enhance the model's recovery by increasing the number of trainable parameters. 
In addition, we applied the same LoRA rank and alpha settings (128 and 256, respectively) to all linear weights in the pruned NLLB model. 
To further enhance the robustness of our experiments, we implemented a LoRA dropout rate of 0.1 across all trials. 

\subsection{Evaluation Metrics}
To evaluate the performance of our models, we employed three key metrics: spBLEU score, number of removing layers, and inference speed. 
The spBLEU score was calculated using the \texttt{SacreBLEU} package with the tokenization of the Flores-200 dataset \cite{costa2022no, papineni-etal-2002-bleu, post-2018-call}. 
Additionally, we measured inference speed by counting the number of tokens generated per second for batch size 1, providing insights into the models' efficiency during deployment. 
Together, these metrics offer a comprehensive view of our models' translation capabilities and operational efficiency.

\subsection{Evaluation Scenarios}
We tested our approach in two distinct scenarios: multi-way translation and single-way translation. 
In the multi-way setting, we aimed to compress the model for translating from German, French, and Persian to English. 
While German and French share some similarities, Persian is notably different, providing a diverse challenge for our method. 
Conversely, in the single-way translation scenario, we focused on compressing the model specifically for translating from English to Persian. 
This approach allowed us to evaluate the model's performance in generating translations for a target language distinct from English.

%

    
    
    


\section{Results}
\label{sec:results}
In this section, we present a comparative analysis of our results for LLaMA3.1-8B-Instruct with other approaches for identifying unimportant layers, focusing on the spBLEU score of pruned models. 
We then compare our final compressed NLLB-3.3B and LLaMA3.1-8B-Instruct models with their original versions in both multi-way and single-way translation tasks, evaluating them in terms of number of pruned layers, token generation speed, and spBLEU score. 
%

\subsection{Comparing Layer Pruning Approaches}

We compared our approach with Top-N and Blockwise pruning methods.
In the Top-N approach, the top N layers of the model are simply removed. 
This method has shown good results in encoder-only language models \cite{voita-etal-2019-bottom, sajjad-etal-2023-layers-effect} and was employed by LLM-Pruner \cite{ma2023llmpruner} in decoder-only layer-wise scenarios. 
In contrast, the Blockwise approach removes a block of layers, typically starting from the penultimate layers \cite{gromov2024unreasonable}. 
We evaluated these methods in a multi-way translation scenario using the LLaMA3.1-8B-Instruct model, fine-tuned for multilingual translation and pruned by 4 layers. 
Although all approaches include a healing phase, no healing was performed in this comparison to focus solely on the identification of unimportant layers.
We specifically compared the number of pruned layers and spBLEU scores across three translation directions: French, German, and Persian to English.

The results, presented in \Cref{tab:comparison}, show that the last layers of the model are crucial, and removing them—or even a block of layers near them—leads to a significant drop in performance. 
In the Top-N approach, layers 28, 29, 30, and 31 were removed, while in the Blockwise approach, layers 25, 26, 27, and 28 were pruned.
However, our approach successfully identified unimportant layers, allowing us to remove layers 10, 11, 19, and 26 while maintaining a spBLEU score close to that of the fine-tuned model.
Note that no healing is performed after pruning.

\begin{table}[t]
  \centering
  \resizebox{0.9\columnwidth}{!}{
  \begin{tabular}{lrrr}
    \toprule
    Approach & Fr$\rightarrow$En & De$\rightarrow$En & Fa$\rightarrow$En \\
    \midrule
    Original-FT & 48.8 & 47.1 & 38.9 \\
    \midrule
    CULL-MT & \textbf{45.8} & \textbf{43.2} & \textbf{33.3} \\
    Top-N & 1.4 & 2.0 & 2.0 \\
    Blockwise & 6.3 & 7.7 & 8.0 \\ 
    \bottomrule
  \end{tabular}
  }
  \caption{Comparison of spBLEU scores after pruning 4 layers using different methods on the LLaMA3.1 model across three translation directions. The first row shows the performance of the original fine-tuned model. Note that no healing is performed after pruning.}
  \label{tab:comparison}
\end{table}

\subsection{Pruning Stopping Threshold}

We ran the layer pruning algorithm without any stopping threshold to compare the results of pruning NLLB-3.3B and LLaMA3.1-8B-Instruct models in the multi-way scenario. 
We calculated the spBLEU score across all directions after pruning each layer and averaged the spBLEU scores. 

\begin{figure}[ht]
  \includegraphics[width=0.95\columnwidth]{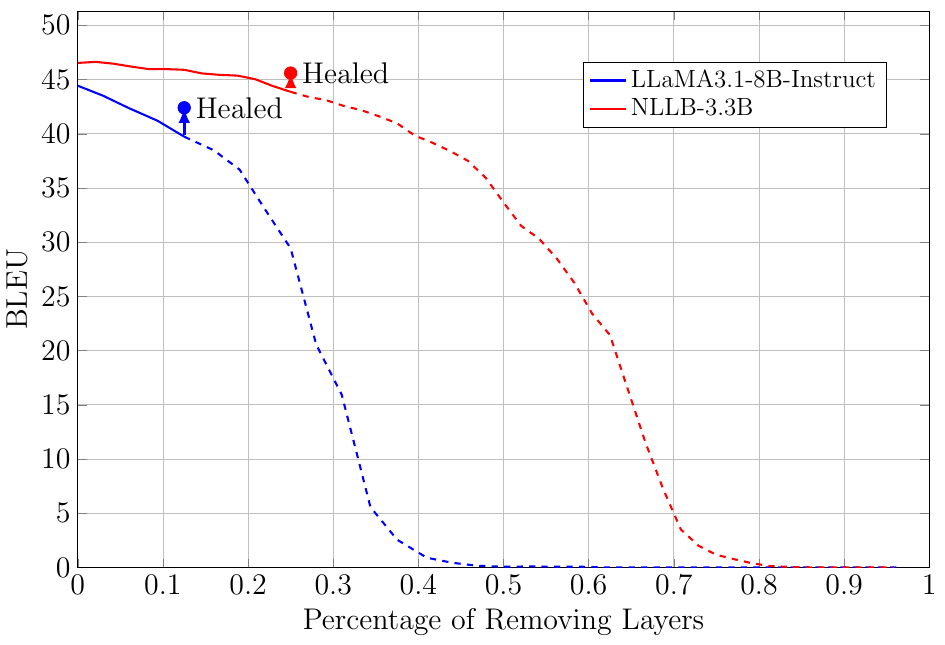}
  \caption{Comparison of layer pruning effects on NLLB-3.3B and LLaMA3.1-8B-Instruct models in a multi-way translation scenario. 
  The NLLB-3.3B model maintains more consistent performance after pruning, whereas the LLaMA3.1-8B-Instruct model shows a steeper performance decline as layers are removed.}
  \label{fig:agg_plot}
\end{figure}

The results, presented in \Cref{fig:agg_plot}, demonstrate that the NLLB-3.3B model, being a natural multilingual NMT model, is more robust against layer pruning. 
In contrast, the fine-tuned LLaMA3.1-8B-Instruct model showed significant sensitivity to the pruning.
After removing 40\% of its layers, the model performance nearly dropped to zero, while the NLLB-3.3B model maintained an average spBLEU score of 40 after losing the same percentage of layers.
As such, we empirically chose the stopping threshold for the NLLB-3.3B model to be a drop of 3.0 in the spBLEU score, while for the LLaMA3.1-8B-Instruct model, it was chosen as 5.0 spBLEU score in both multi-way and single-way scenarios.

\subsection{CULL-MT Evaluation}

We applied our approach, which includes instruction tuning (for the LLaMA3.1-8B-Instruct model), identifying unimportant layers, pruning them, and healing the model through knowledge distillation.

\begin{table*}[h]
  \centering
 \resizebox{0.9\textwidth}{!}{
  \begin{tabular}{llcccccc}
    \toprule
    & & \multicolumn{4}{c}{Multi-way} & \multicolumn{2}{c}{Single-way} \\
    \cmidrule(lr){3-6}
    \cmidrule(lr){7-8}
    Model & Step & De$\rightarrow$En & Fa$\rightarrow$En & Fr$\rightarrow$En & Tok/s & En$\rightarrow$Fa & Tok/s \\
    \midrule
    \multirow{3}{*}{NLLB-3.3B} & Original & 48.4 & 41.5 & 49.7 & 60.9 & 31.8 & 62.7 \\
    & Pruned & 46.1 & 37.8 & 47.7 & 68.7 & 28.7 & 66.6 \\
    & Healed & 47.7 & 39.7 & 49.5 & 68.7 & 30.6 & 66.6 \\
    \midrule
    \multirow{3}{*}{LLaMA3.1-8B-Instruct} & Instruction-Tuned & 46.9 & 38.3 & 48.0 & 35.6 & 26.0 & 40.9 \\
    & Pruned & 41.6 & 30.7 & 43.2 & 43.6 & 17.2 & 46.3 \\
    & Healed & 45.2 & 36.0 & 46.0 & 43.6 & 27.3 & 46.3 \\
    \bottomrule
  \end{tabular}
  }
  \caption{
  spBLEU scores for the NLLB-3.3B and LLaMA3.1-8B-Instruct models across multi-way and single-way translation scenarios. The table includes results for each of the following steps: the original (or instruction-tuned), pruned, and healed models to demonstrate the effectiveness of the healing process. 
  NLLB-3.3B pruned 12 layers for multi-way and 15 for single-way translation, while LLaMA3.1-8B-Instruct pruned 5 and 4 layers, respectively.
  }
  \label{tab:overall}
\end{table*}

The results are presented in \Cref{tab:overall}. 
For NLLB-3.3B, the model demonstrated considerable robustness against layer removal; despite the 3.0 BLEU drop threshold, 12 layers could be pruned in the multi-way scenario and 15 layers in the single-way scenario, representing 25\% and 31\% of the total layers, respectively.

Conversely, the LLaMA3.1-8B-Instruct model was more sensitive to layer pruning, allowing only 5 and 4 layers to be pruned from the overall 32 layers in the multi-way and single-way scenarios, respectively.
Although the performance gap between the pruned and the original (or instruction-tuned) LLaMA3.1-8B-Instruct model was significantly greater than that of the NLLB-3.3B model, the healing process proved to be more effective for LLaMA3.1-8B-Instruct. 
It successfully mitigated much of the spBLEU drop, and in the single-way scenario, the healed model even outperformed the instruction-tuned version.

Additionally, layer-wise pruning had a more noticeable impact on the generation speed (in terms of tokens per second) for the LLaMA3.1-8B-Instruct model. 
As it is a decoder-only architecture, which tends to have slower generation speeds, removing layers noticeably improved its generation speed. 
In contrast, removed layers in the NLLB-3.3B model belong to its encoder, which is a faster module compared to a decoder \cite{vaswani-2017-transformer}.
Thus, pruning these layers did not significantly affect the model’s generation speed.

\begin{figure}[t]
  \includegraphics[width=\columnwidth]{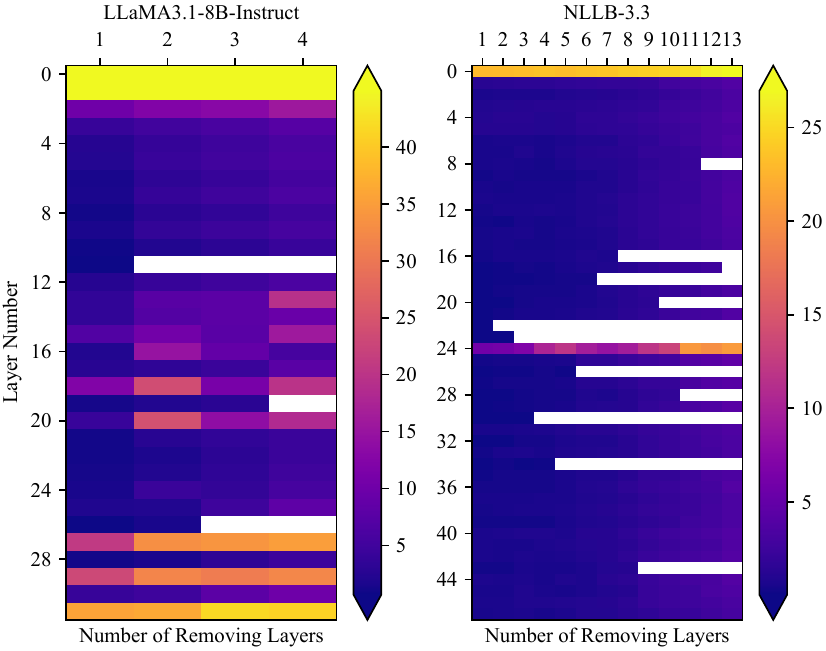}
  \caption{Layer importance during the CULL-MT pruning process, represented by the spBLEU score drop.}
  \label{fig:importance}
\end{figure}

\section{Layer Importance Analysis}
\label{sec:layer_importance}

In this section, we compared the important layers of both models in the multi-way scenario during the process of identifying and removing unimportant layers. 
The results, illustrated in \Cref{fig:importance}, indicate that the color of each layer corresponds to the average spBLEU score drop.
Brighter colors indicate greater spBLEU score drops, meaning the layer is more important, while darker colors signify smaller drops, indicating less importance.
The white color denotes layers that were pruned.

In the NLLB-3.3B model, the first layer of the encoder and the first layer of the decoder were identified as the most important layers. 
The model demonstrated considerable robustness against layer removal, with other layers not becoming crucial when some were pruned. 
Conversely, the LLaMA3.1-8B-Instruct model was highly sensitive to layer pruning. 
The first two layers were critical; removing them caused a significant collapse in performance.
Additionally, the last layers (especially 27, 29, and 31) were also very important, and their removal led to substantial drops in performance.

This sensitivity explains why approaches that remove the top N layers or a block of penultimate layers resulted in large spBLEU score losses.
Notably, while the first and last layers were particularly vital, the middle layers of the LLaMA3.1-8B-Instruct model were important as well; their significance increased with layer removal, suggesting that it is advisable to avoid pruning layers from this region.

\begin{figure}[t]
  \includegraphics[trim={0.8cm 0.8cm 1cm 0.3cm},clip,width=\columnwidth]{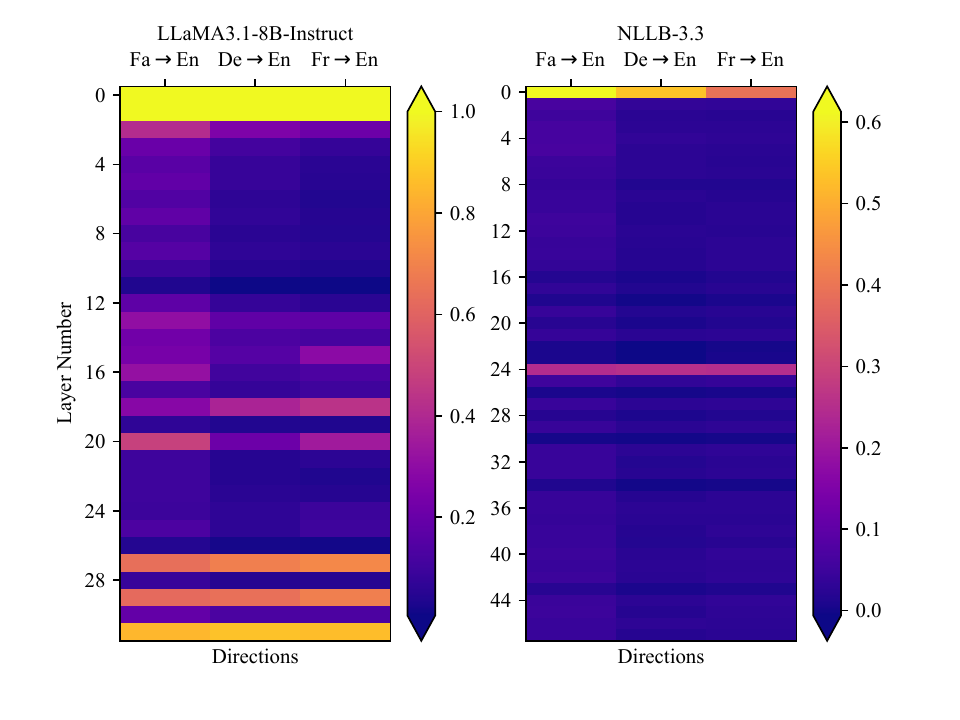}
  \caption{Average layer importance of NLLB-3.3B and LLaMA3.1-8B-Instruct across Fa$\rightarrow$En, Fr$\rightarrow$En, and De$\rightarrow$En translation directions during pruning.}
  \label{fig:direction}
\end{figure}

We also analyzed the importance of layers across different translation directions for both models in the multi-way scenario. 
To accomplish this, we executed our algorithm to prune 4 layers from the LLaMA3.1-8B-Instruct model and 12 layers from the NLLB-3.3B model, calculating the spBLEU score drop for each layer across all directions for each model. 
Since the original models have different baseline scores for each direction, we normalized the dropped spBLEU scores by dividing them by the corresponding baseline values.
This normalization enables comparability of spBLEU scores across different directions.

The results, depicted in \Cref{fig:direction}, indicate that certain layers are generally important across all directions. 
However, we observed that removing the unimportant layers led to a more significant decrease in the spBLEU score for the Persian$\rightarrow$English (Fa$\rightarrow$En) direction compared to the other two directions. 
This demonstrates that the both models are more robust in high-resource directions, while lower-resource directions are more sensitive to layer pruning.

    


\section{Conclusion}
\label{sec:conclusion}
In this paper, we introduced the CULL-MT approach for compressing multilingual NMT models for specific translation directions. 
Our method begins by identifying unimportant layers in a greedy manner and subsequently pruning them. 
To mitigate the damage caused by this pruning, we employed sequence-level knowledge distillation from the original model, utilizing the LoRA fine-tuning technique. 
Our findings revealed that the NLLB-3.3B model demonstrated considerable robustness against layer pruning, while the LLaMA3.1-8B-Instruct model was notably sensitive. 
Nonetheless, our approach effectively pruned both models with minimal performance.

\section{Limitations}
\label{sec:limitations}
The primary limitation of our work is that we tested CULL-MT only on models with fewer than 10 billion parameters due to hardware constraints.
While quantization could enable the application of CULL-MT to larger models, we opted not to use it to avoid introducing additional complexity. 
Consequently, our experiments were conducted on models that fit within our GPU's 24 GB memory.

\bibliography{custom}

\clearpage
\appendix

\section{Dataset size for Healing}
\label{sec:appendix1}

To evaluate the impact of dataset size on the healing process for the NLLB-3.3B model, we conducted experiments focusing on healing the model specifically for the Persian-English translation direction. 
The pruned multiway model was fine-tuned with various dataset sizes generated through sequence-level knowledge distillation from the original model, and evaluated on the NTREX dataset. 
We fine-tuned all parameters of the model for each dataset size and present the results in \Cref{tab:datasize}.

Our findings show that using a very small dataset (10,000 sentences) was insufficient to effectively heal the model's performance drop.
However, a dataset size of 40,000 sentences resulted in recovery of the model's performance.
Moreover, By doubling the dataset size, healing performance showed a slight improvement.
Therefore, to maximize the healed model's performance, we used 160,000 sentences.

\begin{table}[ht]
  \centering
  \resizebox{\columnwidth}{!}{
  \begin{tabular}{cccc}
    \toprule
    Original & Pruned & No. of Sentences & Healed \\
    \midrule
    \multirow{3}{*}{36.2} & \multirow{3}{*}{32.2} & 10,000 & 32.0 \\
    & & 40,000 & 32.8 \\
    & & 80,000 & 33.0 \\
    \bottomrule
  \end{tabular}
  }
  \caption{Evaluating the healing process of the multi-way pruned NLLB-3.3B model with various dataset sizes on the NTREX dataset using the spBLEU score.}
  \label{tab:datasize}
\end{table}

\section{Comparing LoRA with Full-Parameter Fine-tuning}
\label{sec:appendix2}

We conducted an experiment comparing LoRA fine-tuning with full parameter fine-tuning to assess their effectiveness in healing the pruned NLLB-3.3B model. 
For this experiment, we applied sequence-level knowledge distillation using an 80,000 sentence dataset for each translation direction in the multi-way scenario and evaluated our models using the NTREX dataset.

The results, presented in Table \Cref{tab:ft_vs_lora}, indicate that healing the pruned model with the LoRA approach outperforms full parameter fine-tuning.
Therefore, this technique is more suitable for the healing process. 
Additionally, healing the NLLB-3.3B model with LoRA eliminates the need for optimizer state offloading, resulting in faster training times compared to full parameter fine-tuning, which requires optimizer state offloading to function effectively.

\begin{table}[ht]
  \centering
  \resizebox{\columnwidth}{!}{
  \begin{tabular}{lcc}
    \toprule
    Direction & Full-Parameter FT & LoRA \\
    \midrule
    Fa$\rightarrow$En & 33.2 & 34.0 \\
    Fr$\rightarrow$En & 41.2 & 41.6 \\
    De$\rightarrow$En & 42.4 & 43.1 \\
    \midrule
    Average & 38.9 & 39.5 \\
    \bottomrule
  \end{tabular}
  }
  \caption{Comparison of spBLEU scores for healing the multi-way pruned NLLB-3.3B model using the LoRA approach and full parameter fine-tuning.}
  \label{tab:ft_vs_lora}
\end{table}

\end{document}